%% file: main.tex
\documentclass[final]{cvpr}
\usepackage{times}
\usepackage{epsfig}
\usepackage[numbers,sort,compress]{natbib}
\usepackage{graphicx}
\usepackage{amsmath}
\usepackage{amssymb}
\usepackage{here}
\usepackage[pagebackref=true,breaklinks=true,colorlinks,bookmarks=false]{hyperref}
\usepackage{pifont}
\usepackage{cuted}
\usepackage{capt-of}
\usepackage{tabu}
\usepackage{subfig}
\usepackage{soul}
\usepackage{booktabs}
\newcommand{\cmark}{\ding{51}}%

\def\cvprPaperID{5284} 
\def\confYear{CVPR 2021}
\newcommand{\Sref}[1]{Sec.~\ref{#1}}

\newcommand{\Fref}[1]{Fig.~\ref{#1}}
\newcommand{\Tref}[1]{Table~\ref{#1}}

\begin{document}
\input{definitions}

\title{Lipstick ain't enough: Beyond Color Matching for In-the-Wild Makeup Transfer}


\author{
Thao Nguyen$^{1}$ \quad Anh Tuan Tran$^{1,2}$ \quad Minh Hoai$^{1,3}$ \\
$^1$VinAI Research, Hanoi, Vietnam,
$^2$VinUniversity, Hanoi, Vietnam,\\
$^3$Stony Brook University, Stony Brook, NY 11790, USA\\
{\tt\small \{v.thaontp79,v.anhtt152,v.hoainm\}@vinai.io}
}

\makeatletter
\let\@oldmaketitle\@maketitle

\renewcommand{\@maketitle}{\@oldmaketitle
\vspace{-8mm}
\centering
\includegraphics[width=0.95\linewidth,page=5]{imgs/teaser_3.pdf}
\vskip -0.1in
\captionof{figure}{In-the-wild facial makeup consists of both color transfer and pattern addition. We propose a holistic method that can transfer the color and pattern from a reference makeup style to another image.} \label{fig:feature-graphic}

\vspace{3mm}
}
\makeatletter

\maketitle
\begin{abstract}
Makeup transfer is the task of applying on a source face the makeup style from a reference image. Real-life makeups are diverse and wild, which cover not only color-changing but also patterns, such as stickers, blushes, and jewelries. However, existing works overlooked the latter components and confined makeup transfer to color manipulation, focusing only on light makeup styles. In this work, we propose a holistic makeup transfer framework that can handle all the mentioned makeup components. It consists of an improved color transfer branch and a novel pattern transfer branch to learn all makeup properties, including color, shape, texture, and location. To train and evaluate such a system, we also introduce new makeup datasets for real and synthetic extreme makeup. Experimental results show that our framework achieves the state of the art performance on both light and extreme makeup styles. Code is available at \url{https://github.com/VinAIResearch/CPM}.
   
\end{abstract}

\section{Introduction}
\vspace{-2mm}


Across thousands of years of history, humankind has been fascinated with facial beauty. Humans, particularly females, want to be attractive, and facial appearance is a crucial part of this. The cosmetic industry, as reported in 2007, generates a turnover of about \$170 billion each year \cite{cosmeticWiki}. Among face beautification techniques, makeup is the most popular method, accompanied by a wide range of commercial products including foundation, eye shadow, lipsticks, blushes, stickers, facial drawings, and facial accessories.

Due to the popularity of makeup, makeup try-on is an vital application in both retail and entertainment. Among makeup try-on techniques, makeup transfer is the most convenient and effective way. Makeup transfer is the task of transferring the makeup style from one reference face to another face. This task is not trivial; it needs to extract makeup components from the composited reference image. It also needs to analyze the face structure to transfer makeup components between unaligned faces correctly, and there are many factors to account for, including head pose, illumination, facial expressions, and occlusions.




Deep-learning-based generative models are leading methods in tackling this problem. BeautyGAN~\cite{beautygan} and BeautyGlow~\cite{beautyglow} can provide realistic after-makeup images for simple styles on frontal faces. PSGAN \cite{jiang2019psgan} manages to handle faces at various head poses and expressions, while CA-GAN \cite{kips2020cagan} focuses on fine-grained makeup-color matching. However, these methods can only work with simple makeup styles based on color distributed in cosmetic regions such as skin foundations, lipsticks, and eye-shadows. They fail miserably on the complex makeups that rely on shape, texture, and location, such as blushes, face paintings, and makeup jewelries. Only LADN \cite{gu2019ladn} considers these extreme makeups, but its results are far from satisfactory.

In this work, we consider makeup as a combination of color transformation and pattern addition. We aim to transform the color distribution like previous methods while also preserving the shape and appearance of the makeup pattern. To achieve this objective, we introduce a framework with two branches: Color Transfer Branch and Pattern Transfer Branch, which could be run independently in parallel. In the Color Transfer Branch, we employ a CycleGAN-like network structure driven by Histogram Matching as suggested by BeautyGAN~\cite{beautygan}. In the Pattern Transfer Branch, we learn to extract the makeup pattern mask in a supervised manner. Noticeably, unlike previous methods, both our branches work on warped faces in UV space, thus discarding the discrepancy between these faces in terms of shape, head pose, and expression. The results of the two branches are fused to generate the desired output.


We also introduce new makeup-transfer datasets, consisting of both synthetic and real images, and covering a wide range of makeup styles. They include extreme makeup styles, which do not exist in previous makeup datasets. 

\begin{figure*}[t]
\centering
\subfloat[System design]{
  \includegraphics[width=0.64\textwidth,page=3]{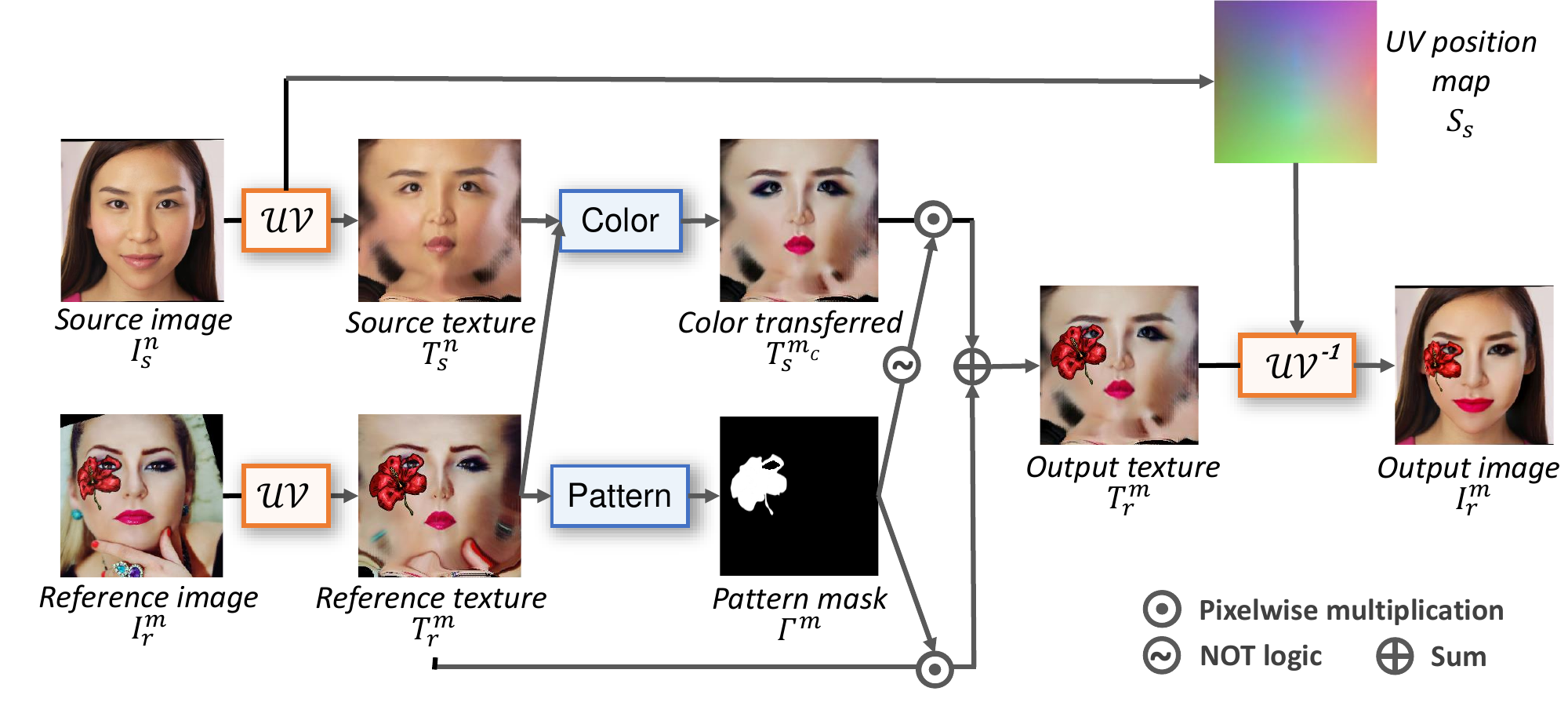}
  \vspace{-5mm}
  \label{fig:overall}
}
~
\subfloat[Comparison with other methods]{
\raisebox{1.5\height} {
\setlength{\tabcolsep}{1.8pt}
\begin{tabu} to .1\textwidth {lccc}
    \hline
    Methods&Color& Control.&Pattern\\ 
    \hline
    BGAN \cite{beautygan}&\cmark&&\\
    BGLOW \cite{beautyglow}&\cmark&\cmark&\\
    LADN \cite{gu2019ladn}&\cmark& &\cmark\\
    PSGAN \cite{jiang2019psgan}&\cmark&\cmark&\\
    CAGAN \cite{kips2020cagan}&\cmark&\cmark&\\
    Ours&\cmark&\cmark&\cmark\\ 
    \hline
\end{tabu}
  \vspace{-5mm}
    \label{tab:comparison_method}
}
}
  \vspace{-2mm}
\caption{Overview of the proposed framework. 'Control.' indicates controllable for partial makeup transfer.}
  \vspace{-5mm}
\end{figure*}

Using the novel network architecture and the newly collected datasets for training, we obtain an all-inclusive makeup transfer method that outperforms all previous methods in terms of coverage, as shown in \Fref{tab:comparison_method}. We also run comprehensive experiments, both qualitative and quantitative, and proposed makeup-transfer benchmarks. Our method outperforms other methods on both light and extreme makeup transfer by a wide margin.


In short, our contributions are: (1) We pose makeup as a combination of color transformation and pattern addition, and develop a comprehensive makeup transfer method that works for both light and extreme styles. (2) We design a novel architecture with two branches for color and pattern transfer, and we propose to use warped faces in the UV space when training two network branches to discard the discrepancy between input faces in terms of shape, head pose, and expression. (3) We introduce new makeup-transfer datasets containing extreme styles that have not been considered in the previous datasets. (4) We obtain state-of-the-art quantitative and qualitative performance.

\section{Related Work} \label{sec:related_makeup}
\vspace{-2mm}
\myheading{Facial Makeup Transfer.} Facial makeup has been studied~\cite{overview_makeup} in computer vision. Given an arbitrary facial image with the desired makeup style, makeup transfer aims to analyze and replicate that makeup to a source image. 

Traditional methods \cite{image_anatogy,example_comestic} focused on image prepossessing techniques, such as landmark extraction and adjustment \cite{makeup_landmark} or reflectance manipulation \cite{makeup_physic}. Recently, due to high-performance hardware and the ability to generate aesthetic images, GANs are widely-used for image-to-image translation tasks, including facial makeup synthesis. CycleGAN-based models \cite{CycleGAN2017,Chang2018PairedCycleGANAS} were introduced to transfer face-to-face makeup styles in an unsupervised manner. For more realistic outcomes, BeautyGAN \cite{beautygan} used Histogram Matching at each facial region to guide the instance-level makeup synthesis. BeautyGLOW \cite{beautyglow} proposed to decompose makeup and non-makeup components in the latent space, using the GLOW framework \cite{glow}. LADN \cite{gu2019ladn} incorporated multiple and overlapping local discriminators for extreme makeup transfers. PSGAN \cite{jiang2019psgan} employed an Attentive Makeup Morphing module to handle transfer across different head poses and facial expressions. Lately, CA-GAN \cite{kips2020cagan} proposed color discriminators to improve fine-grain makeup color transfer at the lips and eye regions.

Most aforementioned methods only consider light makeup based on color transformation in cosmetic regions such as lips and eye-shadows. In-the-wild makeup styles, however, can also cover pattern-based components such as stickers, face drawings, and decoration. To the best of our knowledge, only LADN \cite{gu2019ladn} focused on those extreme makeup styles, however, it has several limitations. First, due to the unsupervised setup, it cannot handle complicated makeup patterns with fine details. Second LADN suffers when the head pose of the source and the reference faces are different, producing noticeable artifacts. Finally, it generates low-quality outputs with evident image degradation traits such as JPEG compression noise and blurry edges.

In this paper, we propose a holistic makeup framework that handles both makeup color and pattern transfer. Our method overcomes the limitations of LADN; it can deal with complicated makeup patterns, be robust to head pose, and produce high-quality outputs.

Although several datasets of makeup faces have been assembled \cite{makeup_in_the_wild,youtube_makeup_dataset,makeup_induced_spoofing_dataset,beautygan,gu2019ladn}, they mainly cover either light or color-focused makeup styles. Since adding patterns is an important part of makeup, pattern-included makeup transfer datasets should be built. We, therefore, introduce such novel datasets for both real and synthetic makeups.





\myheading{3D Face Modelling from a single image.}
To transfer makeup components between faces, we need to understand their facial structures. The human face is a 3D object, and there are many 3D features affecting its appearance in images such as shape, pose, and expression. Thus, reconstructing a 3D face for each input image is crucial to our task.

3D face modeling from single images has been studied for more than two decades \cite{vetter1998estimating}. Among various classical approaches, 3D-morphable models (3DMMs) \cite{blanz1999morphable,paysan09basel,romdhani2003efficient} were the most successful. A 3DMM is a parameterized statistical representation of the 3D faces' manifold learned from 3D face scans. Basel Face Model (BFM) \cite{paysan09basel} is the most popular 3DMM, which approximates any 3D face as a weighted sum of a mean face and principal shape components. The 3D modeling task is then converted to weight optimization so that the composited 3D face is similar to the one in the input image.

Like most other computer vision tasks, 3D face modeling is experiencing fast growth in recent years thanks to deep learning. The first attempts to apply deep learning on this task relied on 3DMM parameter regression via supervised training \cite{Zhu2016Face,tran16_3dmm_cnn,richardson20163d}. MoFA \cite{Tewari_2017_ICCV} proposed to use an auto-encoder for unsupervised training. \citet{tran2018nonlinear} exploited the auto-encoder to learn a nonlinear 3DMM model for better 3D fitting. PRNet \cite{feng2018prn} designed a 2D representation, called UV position map, to encode the aligned 3D face shape, thus converting 3D face modeling to an image-to-image translation problem. The UV representation is easy to use and manipulate, thanks to its 2D form while removing the effect of head poses and expressions. The later 3D face modeling works mainly focused on reconstructing 3D details for realistic modeling \cite{tran2018extreme,Chen_2019_ICCV,yang2020facescape}.

In this paper, we employ PRNet \cite{feng2018prn} in our system since its accuracy is sufficiently good for our task. Furthermore, we take advantage of its UV representation for effective makeup swapping across faces.

\section{Color-\&-Pattern Makeup Transfer Method}
\vspace{-1mm}
Let $I_s^n$ denote the source image ($s$) with non-makeup~($n$) and $I_r^m$ the reference image ($r$) with the desired makeup~($m$). 
Our goal is to obtain $I_s^m$, an image of the source face with transferred makeup from the reference image. It requires learning a function $\mF$ such that: 
\begin{equation}
    \mathcal{F}(I_{s}^{n}, I_{r}^{m})= I_{s}^{m}
\end{equation}

In this section, we will describe our method to learn this function. Our method is designed based on the following two insights. First, the source and target images are not aligned due to different 3D head poses, face shapes, and facial expressions. To remove the misalignment, we should register these images to a uniform template before transferring makeup, and we specifically propose to use the UV map representation. Second, makeup transfer should be viewed as a combination of color transformation and pattern addition. Pattern addition is categorically different from color transformation, and it should be explicitly handled by a specific module of the proposed solution.

In overview, our method consists of the following three steps. First, the input images $I_{s}^{n}, I_{r}^{m}$ are converted to UV texture maps $T_{s}^{n}, T_{r}^{m}$. Second, the texture maps are passed to two parallel branches for color-based and pattern-based makeup transferred. Third, the makeup-transferred texture $T_{s}^{m}$ is formed by combining the outputs of those branches, and this UV texture map is converted to the image space to obtain the final output $I_{s}^{m}$. The pipeline of our method is depicted in \Fref{fig:overall}. In the rest of this section, we will describe the details of the main components. 

\subsection{UV map conversion}\label{sec:uv}
\vspace{-2mm}
UV map representation is a common technique for 3D object texture mapping in computer graphics. The object's texture is flattened into a 2D image, and each 3D vertex of the object is associated with a 2D location on the image, called UV coordinates, for color sampling. PRNet \cite{feng2018prn} extended this idea and introduced a UV position map representation to encode any 3D face shape. It is a 2D image with three channels encoding the XYZ coordinates of the 3D face with respect to the camera coordinates. This UV map is well registered; each pixel in the map corresponds to a fixed semantic point on the face regardless of the input head pose. Alongside the UV position map, we do texture mapping to get the paired texture map. The UV position map packs all information about face shape, head pose, and facial expression, while the mapped texture is invariant to those aspects.

Given an input facial image $I$, we can use the pre-trained model of PRNet, denoted as $\mathcal{UV}$, to extract the corresponding UV position map $S$ and the UV texture $T$. The input image can be recovered from these UV representations via a rendering function $\mathcal{UV}^{-1}$.
\begin{equation}
    S, T := \mathcal{U}\mathcal{V}(I) \quad \textrm{and} \quad I := \mathcal{UV}^{-1}(S, T).
\end{equation}



To transfer makeup between source and reference images with different head poses, we use these UV map presentations. First, we apply the conversion function $\mathcal{U}\mathcal{V}$ on each input image $I_{s}^{n}$ and $I_{r}^{m}$ to get the corresponding UV maps $(S_{s}, T_{s}^{n})$ and $(S_{r}, T_{r}^{m})$. Note that the UV position maps $S_{s}$ and $S_{r}$ depend only on 3D face shapes, thus being independent of the makeup styles. Then, we pass the texture maps $T_{s}^{n}$ and $T_{r}^{m}$ to the color and pattern transfer branches to get makeup swapped in UV space. The outputs of two branches are blended into final texture images $T_{s}^{m}$. Finally, we apply the rendering function to convert it back to standard image representation:
$I_{s}^{m} = \mathcal{UV}^{-1}(S_{s}, T_{s}^{m})$.

\subsection{Color transfer branch}
\label{sec:color_branch}
\vspace{-2mm}
This branch adopts the architecture and training losses proposed in BeautyGAN \cite{beautygan}. The main component is a color-based makeup swapping network $\mathcal{C}$ that swaps makeup color on cosmetic regions between the source and the reference image: $T_{s}^{m_C}, T_{r}^{n_C} :=\mathcal{C}(T_{s}^{n}, T_{r}^{m})$. To train $\mathcal{C}$, it uses a loss function as a weighted sum of the following: 
\begin{itemize} \denselist 
\item \textbf{Adversarial Loss $\mathcal{L}_{adv}$} enforces the output maps $T_{s}^{m_C}$ and $T_{r}^{n_C}$ to be in makeup and non-makeup domain, respectively, using two discriminators,
\item \textbf{Cycle Consistency Loss $\mathcal{L}_{cyc}$} enforces the cycle consistency constraints proposed by CycleGAN \cite{CycleGAN2017},
\item \textbf{Perceptual Loss $\mathcal{L}_{per}$} aims to preserve the identity between the before and after makeup transfer images by using the VGG-16 model pre-trained on ImageNet,
\item \textbf{Histogram Matching Loss $\mathcal{L}_{hist}$} aims to match the color distributions of the reference image and the source image after makeup transfer.
\end{itemize}

The first three loss functions are common, so we omit the detailed discussion here. The final loss, i.e., $\mathcal{L}_{hist}$, is the key loss function proposed by BeautyGAN for transferring makeup color in cosmetic regions. It employs a Histogram Matching ($HM$) function that alters the histogram of the source image to match the reference one in each of several predefined regions: eye shadows, lips, and facial skin. The total loss is a weighted sum of the regional losses:
\begin{equation}
\mathcal{L}_{hist} = \lambda^{eyes}  \mathcal{L}_{hist}^{eyes} + \lambda^{lips}  \mathcal{L}_{hist}^{lips} + \lambda^{skin}  \mathcal{L}_{hist}^{skin},
\end{equation}
where $\lambda^{eyes}, \lambda^{lips}, \lambda^{skin}$ are tunable hyper-parameters.
 
 Each loss term $\mathcal{L}_{hist}^i$ ($i$ can be eyes, lips, or skin) is the distance between the after-makeup image and the histogram-matched version:
\begin{equation}
    \mathcal{L}_{hist}^i = \big|\big|T_{s}^{m_C} \odot \Gamma_{s}^i - HM(T_{s}^{n} \odot \Gamma_{s}^i, T_{r}^{m} \odot \Gamma_{r}^i)\big|\big|.
\end{equation}
where $\odot$ is pixel-wise multiplication, $\Gamma_{s}^i$ and $\Gamma_{r}^i$ are the segmentation masks for region $i$ in the source and the reference image, respectively.

In BeautyGAN, the cosmetic regions are not aligned; they highly differ in size, location, and perspective warping. It severely impacts the histogram match results, reducing the effectiveness of this histogram loss. While being similar to BeautyGAN, our Color Transfer Branch uses the UV texture maps for makeup swapping instead of the original images. This seemingly small innovation actually leads to much improvement. The texture maps are registered pixel-to-pixel, enabling the histogram matching function to work accurately. The region mask is image-invariant and equals to a universal mask: $\Gamma_{s}^i = \Gamma_{r}^i = \Gamma^i$.

We observe that our Color Transfer Branch produces better results compared to BeautyGAN. It captures not only color but also structure and location of the cosmetic makeups, which is crucial to some makeup components such as blushes. We will discuss more this result in \Sref{sec:ablation_uv}.

\subsection{Pattern transfer branch}\label{sec:pattern_transfer}
\label{sec:pattern_branch}
\vspace{-2mm}
Besides the Color Transfer Branch, we propose a novel Pattern Transfer Branch aiming to detect and transfer the pattern-based makeup components such as stickers, facial drawings, and decorative accessories. When transferring these patterns, we need to keep them unchanged in terms of shape, texture, and location but warped to the target 3D surface. In the natural image form, this process is complicated, which includes segmenting the pattern, unwarping it, and re-warping onto the target. Thanks to the UV position map representation, we do not need the unwarping and re-warping steps. The problem reduces to simple image segmentation.


Given the input texture map $T_{r}^{m}$, we aim to extract a binary segmentation mask for its makeup patterns. We can do so by using any segmentation network. In our implementation, a typical UNet structure with a pre-trained Resnet-50 encoder is used. We employ dice loss for training: $\mathcal{L_{DC}} = \frac{2|\Gamma^{gt}\cap \Gamma^{pr}|}{|\Gamma^{gt}|+|\Gamma^{pr}|}$,
where $\Gamma^{gt}$ and $\Gamma^{pr}$ are the ground truth and predicted segmentation masks for the pattern. 

To train this network, we need a makeup dataset with annotated masks for the makeup patterns. However, such datasets do not exist, so we developed ourselves a synthetic dataset, called CPM-Synt-1, for image-mask pair training data. Details of this dataset will be discussed in \Sref{se:dataset_synt1}.



\subsection{Combination}
\vspace{-2mm}
The output of the Pattern Transfer Branch is the pattern mask $\Gamma^{m}$, while the output of the Color Transfer Branch is an entire UV texture map $T_{s}^{m_C}$. The forms of these two outputs are different, reflecting the fundamental differences between two makeup categories. That is why we propose two separate branches for dedicated processing. 

To get the UV texture map for the source image with the desired transferred makeup, we can combine the outputs of the two branches, by blending the reference makeup pattern, defined by the predicted mask $\Gamma^{m}$, with the color-transferred texture map $T_{s}^{m_C}$ from the color transfer branch:
\begin{align}
    & T_{s}^{m} = T_{r}^{m} \odot \Gamma^{m} + T_{s}^{m_C} \odot (1 - \Gamma^{m}).
\end{align}
Finally, we convert this texture map to the output image $I_{s}^{m}$, using the rendering function $I_{s}^{m} = \mathcal{UV}^{-1}(S_{s}, T_{s}^{m})$.

\section{Color-\&-Pattern Makeup (CPM) Datasets}
\vspace{-2mm}

Given the lack of annotated data with extreme makeup styles for the development of in-the-wild makeup transfer methods, we collected this type of data ourselves. In this section, we describe our data collection and generation procedure that led to three \textbf{C}olor and \textbf{P}attern \textbf{M}akeup (CPM) datasets, called CPM-Real, CPM-Synt-1, and CPM-Synt-2.


\subsection{CPM-Real -- In-the-Wild Makeup Dataset}
\vspace{-2mm}

This is a dataset of real faces with real in-the-wild makeups. It is very diverse in terms of makeup styles, containing both color and pattern makeups. The degree of makeup can vary from light to heavy, from color-oriented to pattern-driven. Many images contain extreme makeups, including facial gems, face paintings, hennas, and festival makeups.



To compile this dataset, we first retrieved a set of initial images using keyword searches (e.g., glitter makeup, festival makeup, creative makeup, gems makeup, face painting). We then used the MTCNN face detector \cite{zhang2016joint} to detect and crop faces in each image. We discarded faces smaller than $150{\times}150$. Finally, we manually removed low-quality and inappropriate ones.
The final set has 3895 makeup images, which is 43\% larger than the number of makeup images in~MT~\cite{beautygan}, the previously largest available makeup dataset. This dataset is designed purely for testing purposes.


\begin{figure}[t]
\centering
\vspace{-1.5mm}
\includegraphics[width=.9\linewidth, page=1]{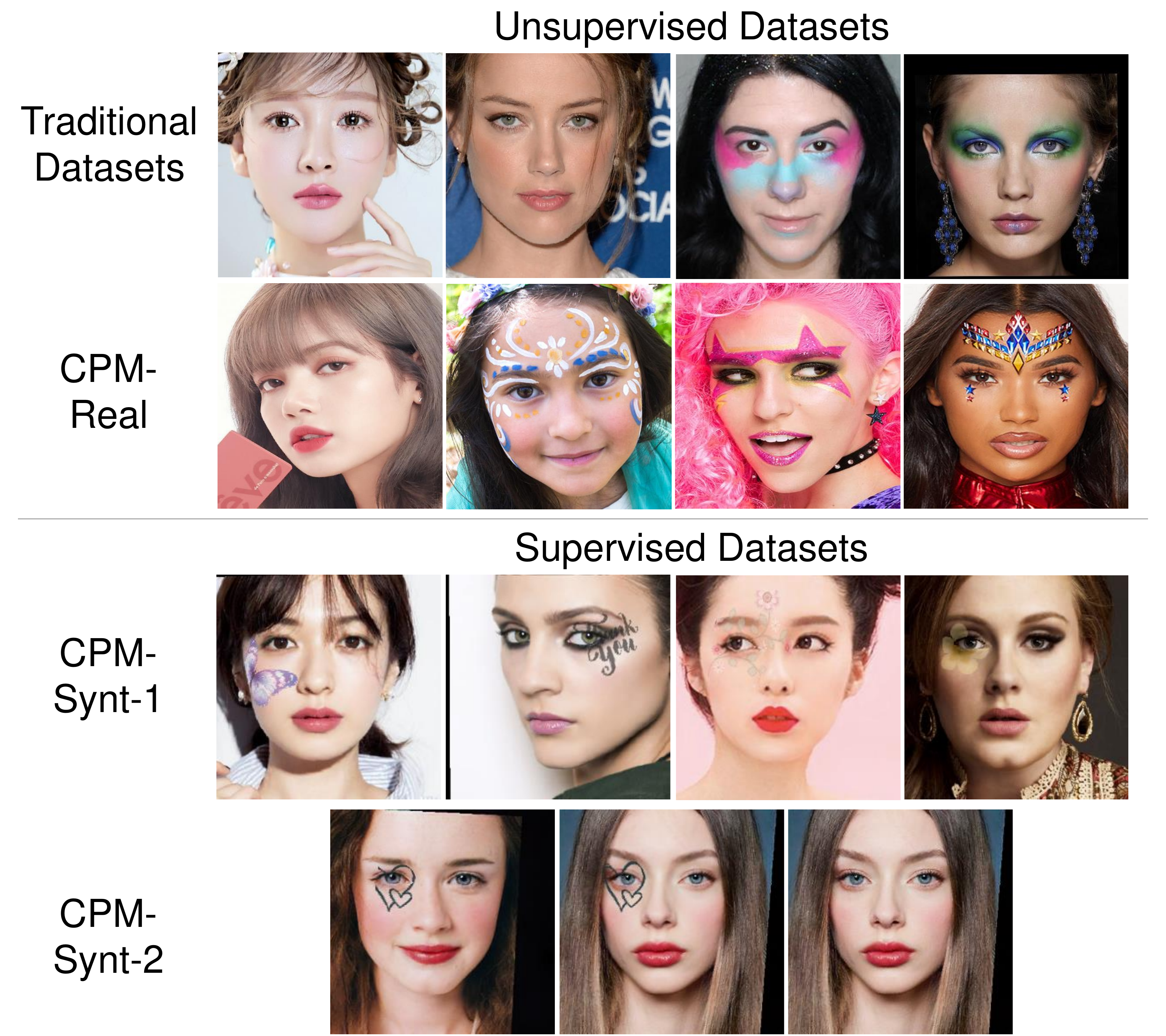}
\vskip -0.1in
  \caption{Features of makeup datasets. First row are from MT \cite{beautygan} and LADN dataset \cite{gu2019ladn}. The rest are from our datasets: CPM-Real, CPM-Synt-1 and CPM-Synt-2.}
    \vspace{-5mm}
\label{fig:makeup_style}
\end{figure}

\begin{figure}[t]
\centering 
\includegraphics[width=0.9\linewidth,page=1]{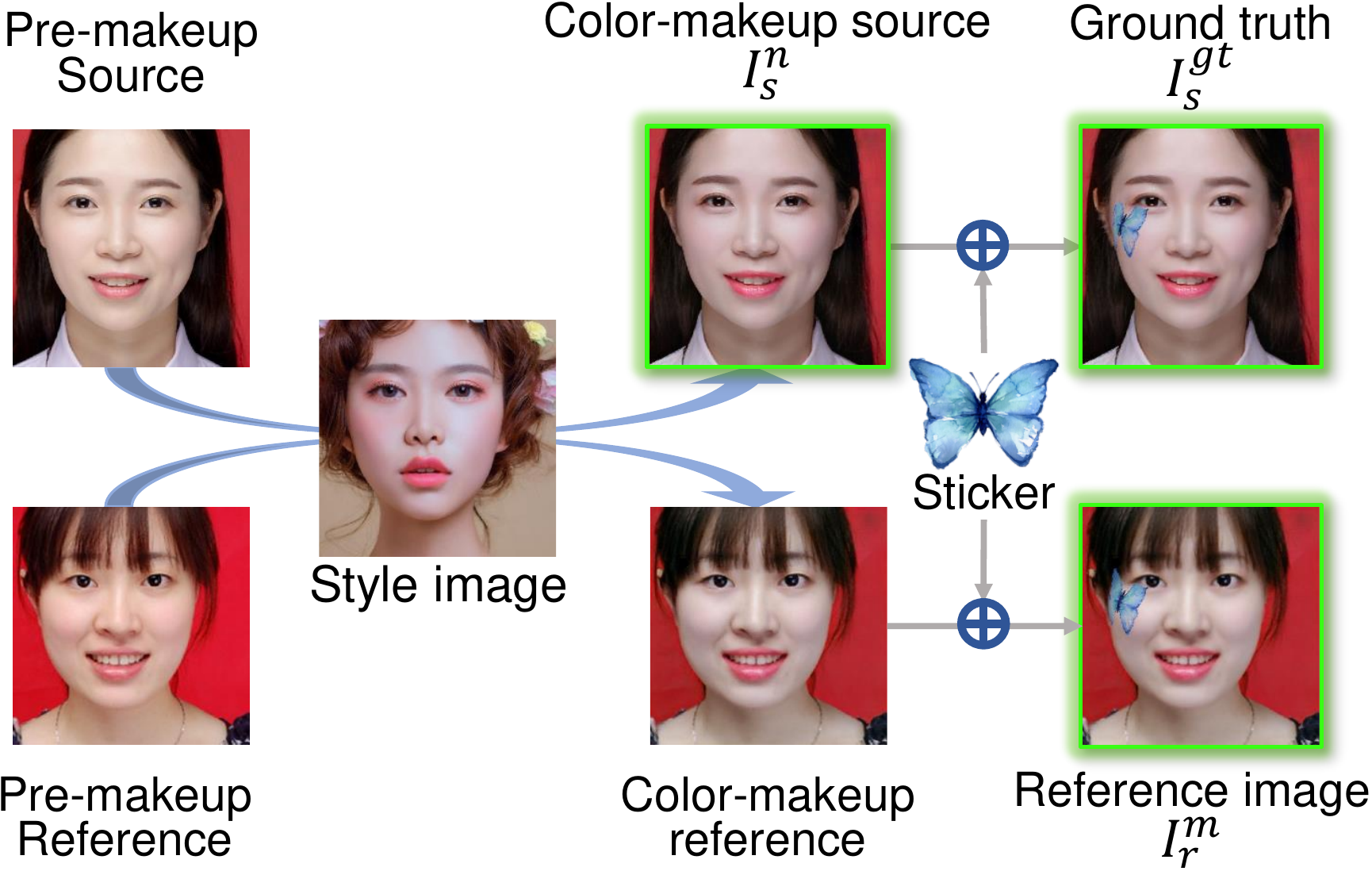}
\vskip -0.1in
   \caption{CPM-Synt-2 synthesizing process}
\label{fig:cpm-synt-2}
\vspace{-3mm}
\end{figure}

\begin{table}[t]
\small
  \centering
\setlength{\tabcolsep}{5.2pt}
  \begin{tabular}[t]{lrccc}
    \toprule
    &\#images&Light&Heavy&Pattern\\
    \hline
    Unsupervised datasets\\
    MT \cite{beautygan}&3834&\cmark&&\\
    LADN \cite{gu2019ladn}&698&\cmark&\cmark&\text{}\\
    M-Wild \cite{jiang2019psgan} &772&\cmark&&\\
    CPM-Real (Ours) &3895&\cmark&\cmark&\cmark\\
    \hline
    Supervised datasets\\
    CPM-Synt-1 (Ours) & 5555&\cmark&\cmark&\cmark\\
    CPM-Synt-2 (Ours) & 1625 &\cmark & &\cmark\\
    \bottomrule 
    \end{tabular}
  \vskip -0.1in
  \caption{Overview of makeup datasets }
    \label{tab:datasets}
    \vspace{-4mm}
\end{table}

\subsection{CPM-Synt-1 -- Added Pattern Dataset}\label{se:dataset_synt1}
\vspace{-1mm}
This is a dataset of real faces with synthetically added makeup patterns. To build it, we needed a set of makeup patterns. Unfortunately, automatic segmentation of makeup patterns from real images was non-trivial, while manual annotation would be laborious due to the tiny little details in many patterns such as henna. To circumvent this problem, we collected some patterns from the Internet with keyword search (e.g., flowers, crystals, gems, henna, daisy, leaf, tattoo) to compile the so-called Stickers dataset with 577 high-quality images. We only used PNG images that had alpha channels, which could later be used for image blending.


Next, we applied the patterns on images from the MT dataset~\cite{beautygan}. As a pattern should conform to the face's surface, we did not directly blend the pattern to the face image. Instead, we applied the blending process in the UV space. The face image was first converted to a UV texture map, as described in Sec.~\ref{sec:uv}. We then blended the pattern on the texture map using its alpha mask with random size, location, and opacity. To make the makeup realistic, we set the pattern's size around the cheek size and put its location inside the face but not at the center. Besides creating the blended texture, we also kept the blending mask as ground-truth for training pattern segmentation module (Sec.~\ref{sec:pattern_transfer}). Finally, we rendered the blended texture to get the after-makeup facial image, together with the sticker segmentation mask. 

In total, CPM-Synt-1 has 5555 after-makeup images. Each image is associated with the ground truth segmentation mask for the pattern and the corresponding UV maps. This dataset split into disjoint training and testing subsets of size 4182 and 1373, respectively. The subjects and the makeup patterns in two subsets are disjointed. 

\subsection{CPM-Synt-2 -- Transferred Pattern Dataset}
\vspace{-1mm}

Despite having ground-truth labels, CPM-Synt-1 does not follow the transfer setup, so it cannot be used to evaluate the pattern-based makeup transfer algorithms. Hence, we built another synthetic dataset called CPM-Synt-2. This dataset contains image triplets: (source image, reference image, ground-truth), specially designed for the pattern-transferred evaluation task. 

One requirement for this test is to have the source and the reference image of the same color-makeup style. Otherwise, we need to impose color-makeup transfer in the ground-truth image. Creating such ground-truth is nontrivial, and no practical solution has been proposed. We can start from non-makeup images, but even these images have a visible cosmetic color difference that requires swapping.

To overcome the mentioned problem, we rely on an assumption of makeup transfer stability: When using the same reference image, a good makeup transfer method will output images of the same makeup style. Based on this assumption, we propose a method to construct the CPM-Synt-2 dataset, as described in Fig. \ref{fig:cpm-synt-2}. First, two non-makeup images are randomly picked from the MT dataset. Then, we transfer both of them to the same makeup style $n$, defined by a Color Style image, using BeautyGAN. This process results in two images with the same color style, called Color-makeup Source $I_s^n$ and Color-makeup Reference $I_r^n$, respectively. Next, we blend the sticker into both images, forming the ground-truth $I_s^{gt}$ and reference image $I_r^m$. Finally, the triplets ($I_s^n, I_r^m, I_s^{gt}$) are formed. \mbox{CPM-Synt-2} consists of 1625 triplets for evaluation purposes.

\section{Experiments}
\subsection{Implementation Details}
We implemented our system with PyTorch. The UV conversation function $\mathcal{UV}$ and the inverse rendering module were based on the existing code and model of PRNet~\cite{feng2018prn}. We trained Pattern Transfer Branch and Color Transfer Branch separately, using respective training datasets.

\myheading{Color Transfer Branch.} For fair comparisons with other methods, we trained $\mathcal{C}$ on the MT dataset \cite{beautygan}. We aligned and resized all images to $256{\times}256$ and then computed their texture maps and facial segmentation in the UV space. The color transfer branch was trained in an unsupervised manner; in each iteration, we randomly sampled one makeup and one non-makeup image to form a swapping pair.

The hyper-parameters were set as follows. The weights for the loss components were: $\lambda_{adv}{=}1$, $\lambda_{cyc}{=}10$, $\lambda_{per}{=}0.005$, and $\lambda_{hist}{=}1$. The weights for histogram matching regions were: $\lambda_{skin}{=}0.1$, $\lambda_{eyes}{=}1$, and $\lambda_{lips}{=}1$. Batch size was set to one. We used Adam optimizer with learning rate 0.0002 to train the network until convergence.

\myheading{Pattern Transfer Branch.} This branch was trained with supervised learning, using the CPM-Synt-1 dataset. Each training image came with the pattern segmentation mask, both having size $256{\times}256$. We utilized UNet structure with Resnet-50 as the pre-trained encoder. Since the original segmentation mask was non-binary, we used sigmoid as activation function. The model was trained for 300 epochs with batch size~8, Adam optimizer, and learning rate 0.0001.


\begin{figure*}[t]
\centering 
\includegraphics[width=.85\textwidth,page=2]{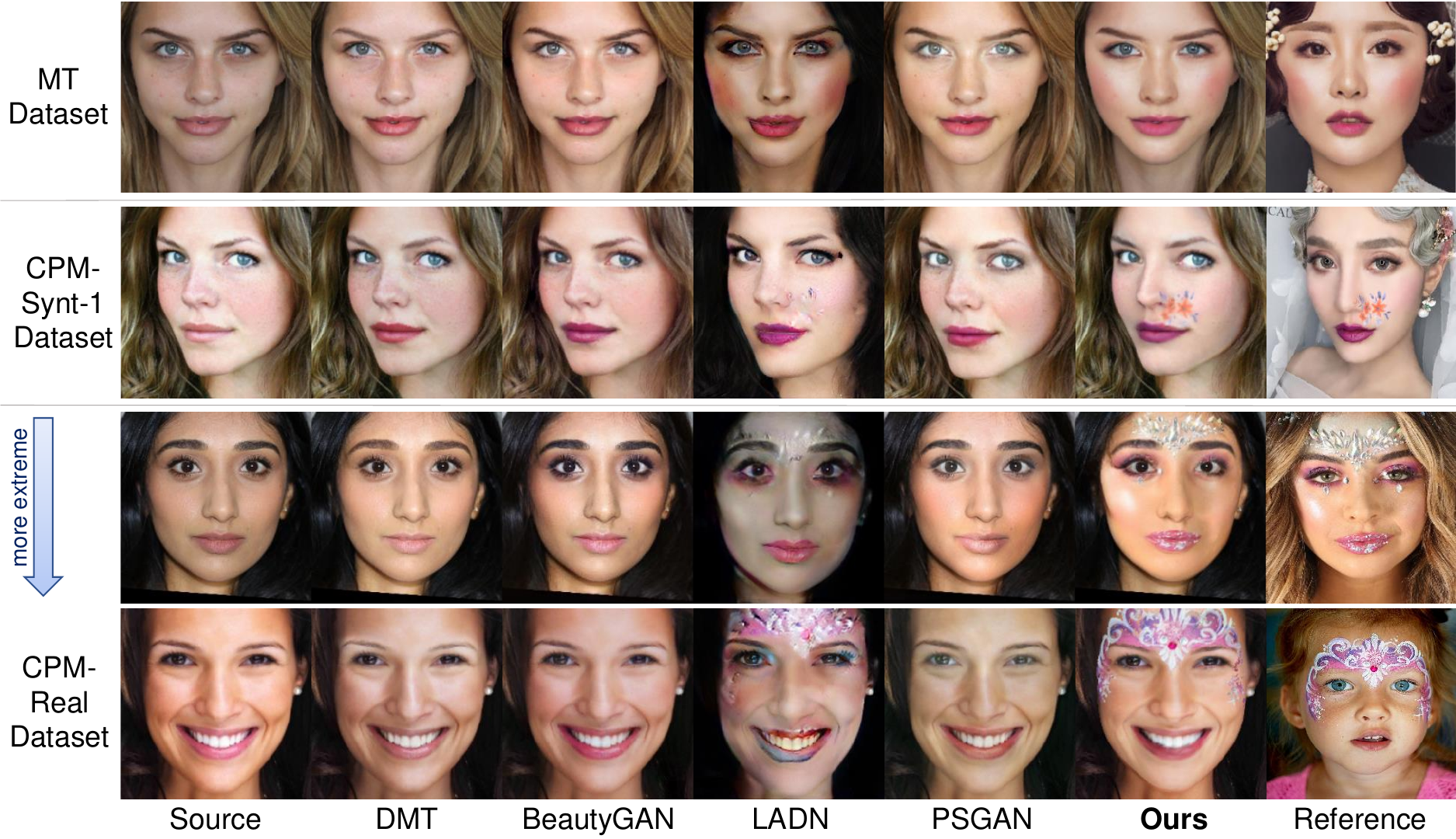}
\vskip -0.1in
   \caption{Qualitative Results}
   \vspace{-3mm}
\label{fig:qualitative_result}
\end{figure*}

\subsection{Qualitative experiments}
\vspace{-2mm}
We compared the proposed method to the state-of-the-art methods, including DMT \cite{dmt}, BeautyGAN \cite{beautygan}, LADN \cite{gu2019ladn}, and PSGAN \cite{jiang2019psgan}. We skipped some baselines, such as BeautyGlow \cite{beautyglow} and CA-GAN \cite{kips2020cagan} because they are both color-only and have no released model. The evaluations were conducted on both the existing and proposed datasets. We present here a few qualitative results but more examples can be found in the supplementary.

\myheading{MT dataset.} We conducted an experiment on the existing MT dataset \cite{beautygan} to examine the ability to transfer color-based makeup styles. As can be seen in the first row of \Fref{fig:qualitative_result}, our model can capture well the lips' color, similar to the state-of-the-art BeautyGAN \cite{beautygan}. Moreover, thanks to UV-based swapping solution, our method can successfully transfer the face blushes and is the only method that captures the glowing skin foundation. 


\myheading{CPM-Synt-1 dataset.} We evaluated the transfer results  with the presence of synthesized makeup patterns. For each reference makeup in the test set of CPM-Synt-1, we randomly picked a source image in the MT dataset and do makeup transfer. A representative result is shown in the second row of \Fref{fig:qualitative_result}. Although the makeup pattern was unseen during training, our network could capture its pattern well and transport it to the output. All other methods, including LADN, failed to handle such a complicated style.

\myheading{CPM-Real dataset.} Finally, we tested with real in-the-wild makeup styles. This time, we used the reference makeup in the CPM-Real dataset, while the source image was still from the MT dataset. We present two examples in the last rows of \Fref{fig:qualitative_result}. Although providing realistic results, color-based methods completely ignored facial drawings and decoration. LADN could partially replicate the reference styles, but its results are unnatural and unappealing. Our method could retain the makeup pattern details and return the results that are closest to desirable makeups.


\begin{table}[t]
\small
  \centering
\setlength{\tabcolsep}{2.5pt}
  \begin{tabular}[t]{lccccc}
    \toprule
    Dataset & DMT& BeautyGAN& LADN& PSGAN& Ours \\ \midrule
    MT \cite{beautygan} & 2.99& 3.69& 2.39& 3.25& \textbf{4.35} \\
    CPM-Synt-1& 3.28& 3.50& 1.67& 2.92& \textbf{4.89} \\
    CPM-Real& 2.50& 2.87& 2.24 & 2.76& \textbf{4.60} \\ \bottomrule 
  \end{tabular}
  \vskip -0.1in
  \caption{{\bf User survey results} for the qualitative results from three datasets. The numbers shown are the average user ratings, with $5$ being the perfect score and $1$ the lowest. Our method achieves the highest scores on all three surveys.}
    \label{tab:user_survey}
\end{table}

\begin{table}[t]
\small
  \centering
\setlength{\tabcolsep}{3pt}
  \begin{tabular}[t]{llccccc}
    \toprule
    Dataset & Metric & DMT& BGAN& LADN& PSGAN& Ours \\ \midrule
    Synt-1 & mIOU & -& -& -& -& \textbf{0.788} \\ \midrule
    Synt-2 & MS-SSIM &0.918& 0.918& 0.656& 0.723& \textbf{0.977} \\\bottomrule 
  \end{tabular}
  \vskip -0.1in
  \caption{\textbf{Ground-truth experiments:} Pattern segmentation on CPM-Synt-1 (top row) and Makeup transfer on CPM-Synt-2 (bottom row).}
    \label{tab:segmentation_perf}
    \vspace{-3mm}
\end{table}

\subsection{User surveys}
For the subjective evaluation of the results, we conducted a user survey for each dataset above. Each survey consisted of 20 questions. For each question, the participants were asked to rank the after-makeup images from the best to the worst. Subsequently, we assigned a score of 5 to the highest-ranked method, and 1 to the lowest one. There were 40 participants, leading to 800 answers for each survey. 


The average survey scores are reported in \Tref{tab:user_survey}. Our method outperforms the others by a wide margin on all tests. Its scores are close to the perfect score of 5, suggesting the superiority of our method in almost all questions. 

\begin{figure*}[t]
\centering 
\includegraphics[width=0.9\linewidth,page=1]{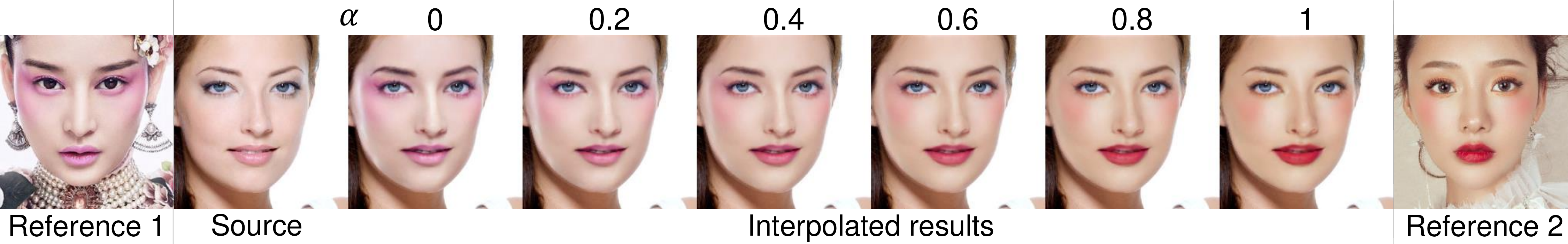}
\vskip -0.1in
\caption{{\bf Makeup style interpolation}. The middle images have makeup style interpolated from two reference styles with a mixing parameter $\alpha$.}
\label{fig:interpolation}
\vspace{-5mm}
\end{figure*}

\subsection{Ground-truth experiments}
\vspace{-1mm}
By building labelled datasets, we can conduct quantitative experiments that have not been done in the previous studies. We first compute our pattern segmentation network's accuracy, then evaluate the quality of the makeup-transfer results produced by ours and baseline methods.

\myheading{Makeup pattern segmentation.}
We evaluated the performance of our pattern segmentation branch on the test set of CPM-Synt-1, and the result is shown in the first row of \Tref{tab:segmentation_perf}. Our pattern segmentation branch achieved 0.788 mIoU. It is not perfect, but sufficiently good for the down-stream task of makeup transfer. 



\myheading{Makeup transfer quality.}
To quantitatively compare our method and other baselines in the makeup-transfer setting, we conducted experiments on the CPM-Synt-2 dataset. We used the MS-SSIM metric to evaluate the quality of the after-makeup images in comparison with ground-truth ones. The average score for each method is reported in the second row of \Tref{tab:segmentation_perf}. The MS-SSIM of our method is 0.977, surpassing the second method by a wide margin.

\subsection{Ablation Studies}
\label{sec:ablation_studies}
\begin{figure}
\centering 
\includegraphics[width=.85\linewidth,page=1]{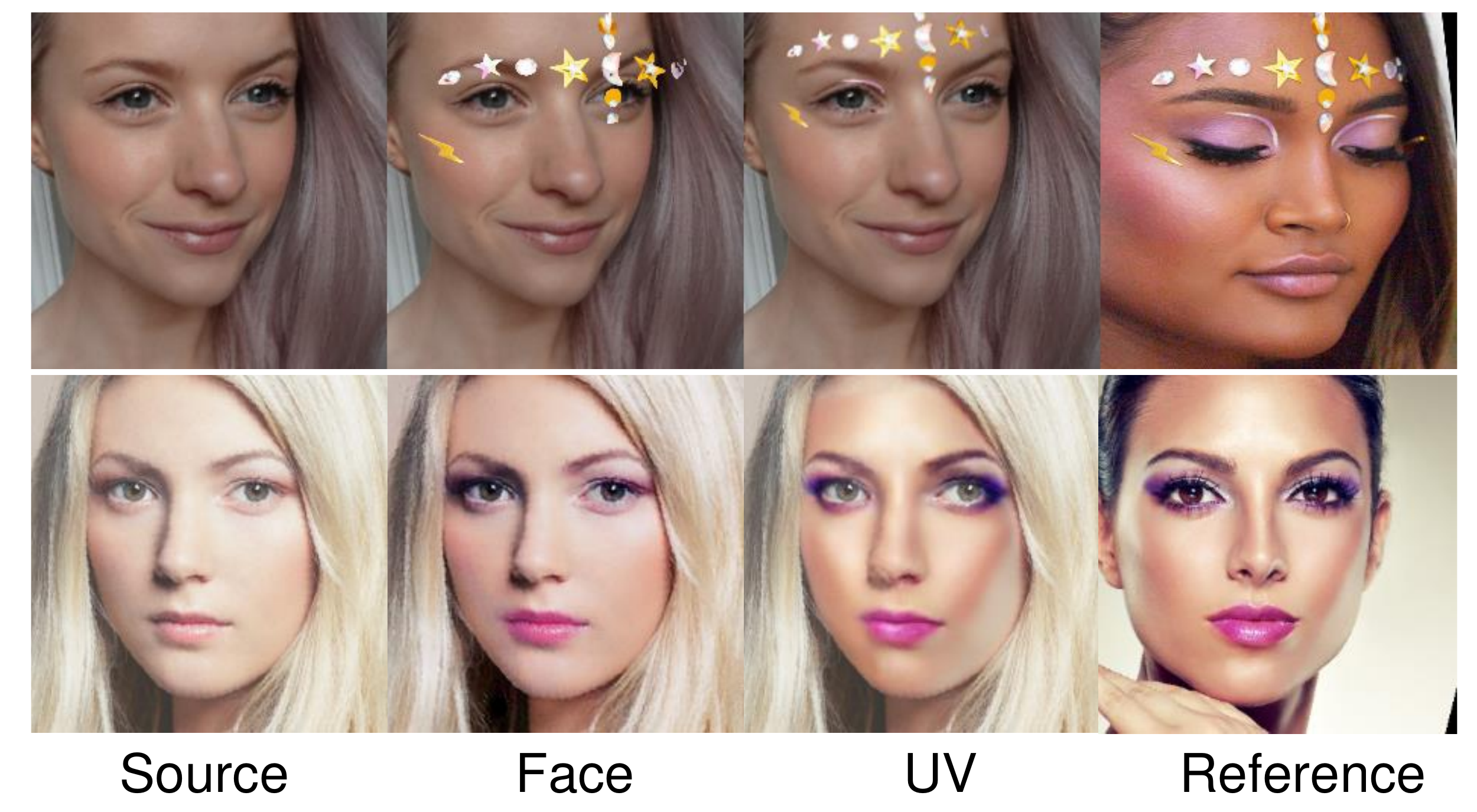}
\vskip -0.1in
  \caption{{\bf Benefits of the UV space} to Pattern Transfer Branch (first row) and Color Transfer Branch (second row). From left to right: Source image, results obtained by training on the original image space, results obtained by training on the UV space, and the reference image.}
\label{fig:texture}
\vspace{-2mm}
\end{figure}

\vspace{-2mm}
\myheading{UV-based makeup transfer.}\label{sec:ablation_uv}
As discussed in \Sref{sec:pattern_transfer}, the UV representation is critical to the Pattern Branch. \Fref{fig:texture} shows in the first row a comparison between pattern in image space and in UV space. Our method aligns the source and target faces pixel-by-pixel, removing the differences in 3D poses, shapes, and expressions. Hence, we can transfer the pattern easily and precisely.

Last row of \Fref{fig:texture} compares the makeup transfer results between BeautyGAN, trained on original faces, and our Color Transfer Branch, trained on the UV space. As can be seen, our method replicates the reference style much more accurately. It preserves both the purple eye shadow and the glowing skin foundation.


\myheading{Identity preservation.}  We used ArcFace \cite{deng2018arcface} to calculate the similarity score between the faces before and after makeup transfer. The average similarity scores on the MT and CPM-Synt-1 datasets are 0.851 and 0.781, respectively. Based on the recommended face verification threshold of 0.5, our makeup transfer method preserves the identities of the subjects. These similarity scores are lower than the maximum score of 1, but this is expected because real-life facial makeup may also change facial characteristics dramatically.

\begin{figure}[t]
\centering 
\includegraphics[width=.85\linewidth,page=1]{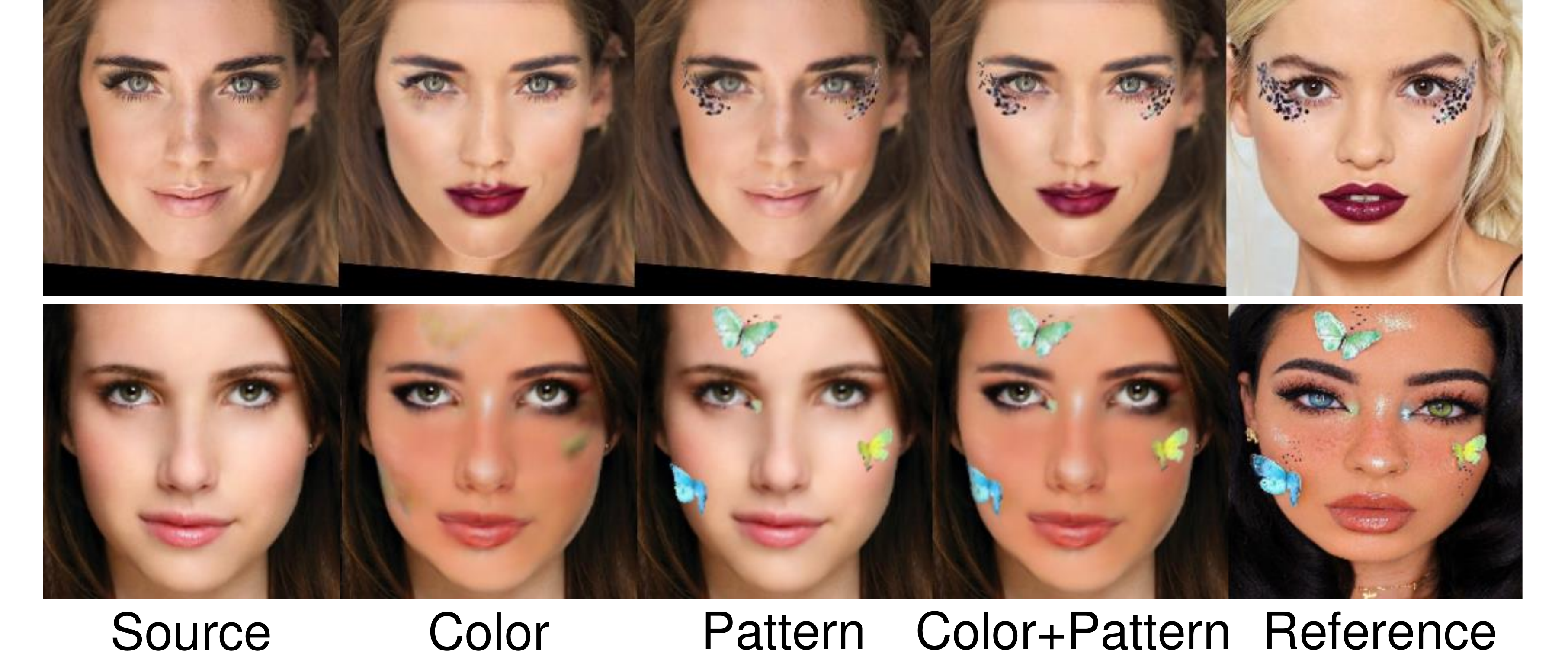}
\vskip -0.1in
   \caption{Branch Analysis.}
\label{fig:ablation_study_branch}
\vspace{-5mm}
\end{figure}

\myheading{Branch Analysis.}
Both color and pattern branches are vital, as illustrated in  \Fref{fig:ablation_study_branch}. The Color Transfer Branch alone failed to bring the face drawings and stickers from the reference image to the target face. When using Pattern Branch only, the lip color of the output image stays the same as the original. We need to combine two branches to replicate all makeup components of the reference image.

\begin{figure}[t]
\centering
\includegraphics[width=.95\linewidth,page=1]{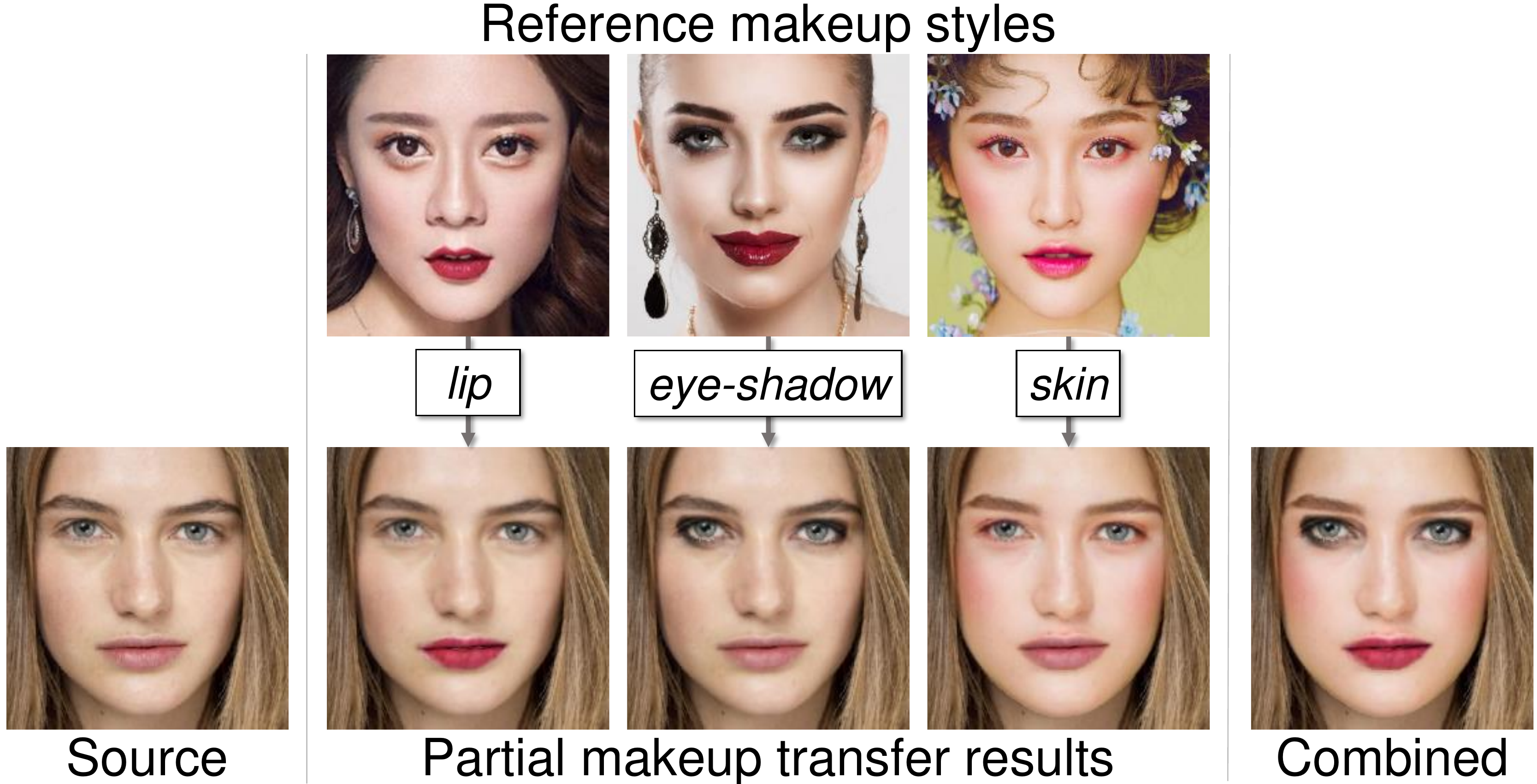}
\vskip -0.1in
   \caption{{\bf Partial makeup transfer}. First rows are different styles. Second rows are results from partial makeup transfer only (lip, eye-shadow, skin), and combination of all styles.}
\label{fig:partial}
\vspace{-5mm}
\end{figure}
\subsection{Interpolation and Partial Makeup Transfer}
\vspace{-2mm}
\myheading{Interpolation.} Makeup interpolation is an interesting application of makeup transfer. While interpolating between makeup patterns is not practical, interpolating between makeup colors is pretty common and easy. Given a single input image $I_s^n$ and two reference styles $I_{r_1}^{m_1}$ and $I_{r_2}^{m_2}$, we can run two makeup transfer processes in parallel to get the color-transferred texture maps $T_{s}^{m_1}$ and $T_{s}^{m_2}$. We can then mix these texture maps by a mixing parameter $\alpha \in [0, 1]$, and render to get the interpolated output. \Fref{fig:interpolation} displays some interpolated results in case one or two reference styles are given. The results are smooth and natural, even in extreme regions such as heavy eye-shadow and cheek color.

\myheading{Partial makeup transfer.} Further exploiting the UV position map, we can use it together with facial segmentation to perform partial makeup transfer. Instead of transferring makeup on the entire face, we can do it on a face region defined by some input mask. This controllable mechanism was proposed in the previous works \cite{beautyglow,jiang2019psgan} and can be easily implemented in our system. \Fref{fig:partial} provides an example in which we transferred makeup partially for the lips, eye shadow, and skin region, then generated a makeup composition on the entire face. 


\section{Conclusion}
\vspace{-2mm}
In this paper, we extend the definition of the makeup transfer task and propose a novel holistic framework to deal with in-the-wild makeup styles. Makeup styles are now interpreted as a combination of color-matching and pattern-addition, respectively, solved by our Color Transfer Branch and Pattern Transfer Branch. UV representation is incorporated to improve the results of both branches. The experiments show our framework can achieve state-of-the-art qualitative and quantitative results. Moreover, we propose novel datasets to leverage  makeup-transfer studies and encourage future development.

{\small
\setlength{\bibsep}{0pt}
\bibliographystyle{plainnat}
\setlength{\bibsep}{0pt}
\bibliography{longstrings,egbib}
}

\end{document}

%% file: definitions.tex
\def\mA{\mathcal{A}}
\def\mB{\mathcal{B}}
\def\mC{\mathcal{C}}
\def\mD{\mathcal{D}}
\def\mE{\mathcal{E}}
\def\mF{\mathcal{F}}
\def\mG{\mathcal{G}}
\def\mH{\mathcal{H}}
\def\mI{\mathcal{I}}
\def\mJ{\mathcal{J}}
\def\mK{\mathcal{K}}
\def\mL{\mathcal{L}}
\def\mM{\mathcal{M}}
\def\mN{\mathcal{N}}
\def\mO{\mathcal{O}}
\def\mP{\mathcal{P}}
\def\mQ{\mathcal{Q}}
\def\mR{\mathcal{R}}
\def\mS{\mathcal{S}}
\def\mT{\mathcal{T}}
\def\mU{\mathcal{U}}
\def\mV{\mathcal{V}}
\def\mW{\mathcal{W}}
\def\mX{\mathcal{X}}
\def\mY{\mathcal{Y}}
\def\mZ{\mathcal{Z}} 

\def\bbN{\mathbb{N}} 
\def\bbR{\mathbb{R}} 
\def\bbP{\mathbb{P}} 
\def\bbQ{\mathbb{Q}} 
\def\bbE{\mathbb{E}}

\def\1n{\mathbf{1}_n}
\def\0{\mathbf{0}}
\def\1{\mathbf{1}}

\def\A{{\bf A}}
\def\B{{\bf B}}
\def\C{{\bf C}}
\def\D{{\bf D}}
\def\E{{\bf E}}
\def\F{{\bf F}}
\def\G{{\bf G}}
\def\H{{\bf H}}
\def\I{{\bf I}}
\def\J{{\bf J}}
\def\K{{\bf K}}
\def\L{{\bf L}}
\def\M{{\bf M}}
\def\N{{\bf N}}
\def\O{{\bf O}}
\def\P{{\bf P}}
\def\Q{{\bf Q}}
\def\R{{\bf R}}
\def\S{{\bf S}}
\def\T{{\bf T}}
\def\U{{\bf U}}
\def\V{{\bf V}}
\def\W{{\bf W}}
\def\X{{\bf X}}
\def\Y{{\bf Y}}
\def\Z{{\bf Z}}

\def\a{{\bf a}}
\def\b{{\bf b}}
\def\c{{\bf c}}
\def\d{{\bf d}}
\def\e{{\bf e}}
\def\f{{\bf f}}
\def\g{{\bf g}}
\def\h{{\bf h}}
\def\i{{\bf i}}
\def\j{{\bf j}}
\def\k{{\bf k}}
\def\l{{\bf l}}
\def\m{{\bf m}}
\def\n{{\bf n}}
\def\o{{\bf o}}
\def\p{{\bf p}}
\def\q{{\bf q}}
\def\r{{\bf r}}
\def\s{{\bf s}}
\def\t{{\bf t}}
\def\u{{\bf u}}
\def\v{{\bf v}}
\def\w{{\bf w}}
\def\x{{\bf x}}
\def\y{{\bf y}}
\def\z{{\bf z}}

\def\balpha{\mbox{\boldmath{$\alpha$}}}
\def\bbeta{\mbox{\boldmath{$\beta$}}}
\def\bdelta{\mbox{\boldmath{$\delta$}}}
\def\bgamma{\mbox{\boldmath{$\gamma$}}}
\def\blambda{\mbox{\boldmath{$\lambda$}}}
\def\bsigma{\mbox{\boldmath{$\sigma$}}}
\def\btheta{\mbox{\boldmath{$\theta$}}}
\def\bomega{\mbox{\boldmath{$\omega$}}}
\def\bxi{\mbox{\boldmath{$\xi$}}}
\def\bnu{\mbox{\boldmath{$\nu$}}}                                  
\def\bphi{\mbox{\boldmath{$\phi$}}}
\def\bmu{\mbox{\boldmath{$\mu$}}}

\def\bDelta{\mbox{\boldmath{$\Delta$}}}
\def\bOmega{\mbox{\boldmath{$\Omega$}}}
\def\bPhi{\mbox{\boldmath{$\Phi$}}}
\def\bLambda{\mbox{\boldmath{$\Lambda$}}}
\def\bSigma{\mbox{\boldmath{$\Sigma$}}}
\def\bGamma{\mbox{\boldmath{$\Gamma$}}}
                                  
\newcommand{\myprob}[1]{\mathop{\mathbb{P}}_{#1}}

\newcommand{\myexp}[1]{\mathop{\mathbb{E}}_{#1}}

\newcommand{\mydelta}[1]{1_{#1}}

\newcommand{\myminimum}[1]{\mathop{\textrm{minimum}}_{#1}}
\newcommand{\mymaximum}[1]{\mathop{\textrm{maximum}}_{#1}}    
\newcommand{\mymin}[1]{\mathop{\textrm{minimize}}_{#1}}
\newcommand{\mymax}[1]{\mathop{\textrm{maximize}}_{#1}}
\newcommand{\mymins}[1]{\mathop{\textrm{min.}}_{#1}}
\newcommand{\mymaxs}[1]{\mathop{\textrm{max.}}_{#1}}  
\newcommand{\myargmin}[1]{\mathop{\textrm{argmin}}_{#1}} 
\newcommand{\myargmax}[1]{\mathop{\textrm{argmax}}_{#1}} 
\newcommand{\myst}{\textrm{s.t. }}

\newcommand{\denselist}{\itemsep -1pt}
\newcommand{\sparselist}{\itemsep 1pt}

\definecolor{pink}{rgb}{0.9,0.5,0.5}
\definecolor{purple}{rgb}{0.5, 0.4, 0.8}   
\definecolor{gray}{rgb}{0.3, 0.3, 0.3}
\definecolor{mygreen}{rgb}{0.2, 0.6, 0.2}

\newcommand{\cyan}[1]{\textcolor{cyan}{#1}}
\newcommand{\red}[1]{\textcolor{red}{#1}}  
\newcommand{\blue}[1]{\textcolor{blue}{#1}}
\newcommand{\magenta}[1]{\textcolor{magenta}{#1}}
\newcommand{\pink}[1]{\textcolor{pink}{#1}}
\newcommand{\green}[1]{\textcolor{green}{#1}} 
\newcommand{\gray}[1]{\textcolor{gray}{#1}}    
\newcommand{\mygreen}[1]{\textcolor{mygreen}{#1}}    
\newcommand{\purple}[1]{\textcolor{purple}{#1}}       

\definecolor{greena}{rgb}{0.4, 0.5, 0.1}
\newcommand{\greena}[1]{\textcolor{greena}{#1}}

\definecolor{bluea}{rgb}{0, 0.4, 0.6}
\newcommand{\bluea}[1]{\textcolor{bluea}{#1}}
\definecolor{reda}{rgb}{0.6, 0.2, 0.1}
\newcommand{\reda}[1]{\textcolor{reda}{#1}}

\def\changemargin#1#2{\list{}{\rightmargin#2\leftmargin#1}\item[]}
\let\endchangemargin=\endlist
                                               
\newcommand{\cm}[1]{}

\newcommand{\mhoai}[1]{{\color{magenta}\textbf{[Hoai: #1]}}}

\newcommand{\mtodo}[1]{{\color{red}$\blacksquare$\textbf{[TODO: #1]}}}
\newcommand{\myheading}[1]{\vspace{1ex}\noindent \textbf{#1}}
\newcommand{\htimesw}[2]{\mbox{$#1$$\times$$#2$}}


\newif\ifshowsolution
\showsolutiontrue

\ifshowsolution  
\newcommand{\Comment}[1]{\paragraph{\bf $\bigstar $ COMMENT:} {\sf #1} \bigskip}
\newcommand{\Solution}[2]{\paragraph{\bf $\bigstar $ SOLUTION:} {\sf #2} }
\newcommand{\Mistake}[2]{\paragraph{\bf $\blacksquare$ COMMON MISTAKE #1:} {\sf #2} \bigskip}
\else
\newcommand{\Solution}[2]{\vspace{#1}}
\fi

\newcommand{\truefalse}{
\begin{enumerate}
	\item True
	\item False
\end{enumerate}
}

\newcommand{\yesno}{
\begin{enumerate}
	\item Yes
	\item No
\end{enumerate}
}


%% file: main.bbl
\begin{thebibliography}{33}
\providecommand{\natexlab}[1]{#1}
\providecommand{\url}[1]{\texttt{#1}}
\expandafter\ifx\csname urlstyle\endcsname\relax
  \providecommand{\doi}[1]{doi: #1}\else
  \providecommand{\doi}{doi: \begingroup \urlstyle{rm}\Url}\fi

\bibitem[Blanz and Vetter(1999)]{blanz1999morphable}
V.~Blanz and T.~Vetter.
\newblock Morphable model for the synthesis of {3D} faces.
\newblock In \emph{Proceedings of the ACM SIGGRAPH Conference on Computer
  Graphics}, 1999.

\bibitem[Chang et~al.(2018)Chang, Lu, Yu, and
  Finkelstein]{Chang2018PairedCycleGANAS}
Huiwen Chang, Jingwan Lu, Fisher Yu, and Adam Finkelstein.
\newblock Pairedcyclegan: Asymmetric style transfer for applying and removing
  makeup.
\newblock \emph{Proceedings of the {IEEE} Conference on Computer Vision and
  Pattern Recognition}, 2018.

\bibitem[Chen et~al.(2019{\natexlab{a}})Chen, Chen, Zhang, Mitchell, and
  Yu]{Chen_2019_ICCV}
Anpei Chen, Zhang Chen, Guli Zhang, Kenny Mitchell, and Jingyi Yu.
\newblock Photo-realistic facial details synthesis from single image.
\newblock In \emph{Proceedings of the International Conference on Computer
  Vision}, 2019{\natexlab{a}}.

\bibitem[{Chen} et~al.(2013){Chen}, {Dantcheva}, and
  {Ross}]{makeup_in_the_wild}
C.~{Chen}, A.~{Dantcheva}, and A.~{Ross}.
\newblock Automatic facial makeup detection with application in face
  recognition.
\newblock In \emph{International Conference on Biometrics}, 2013.

\bibitem[{Chen} et~al.(2017){Chen}, {Dantcheva}, {Swearingen}, and
  {Ross}]{makeup_induced_spoofing_dataset}
C.~{Chen}, A.~{Dantcheva}, T.~{Swearingen}, and A.~{Ross}.
\newblock Spoofing faces using makeup: An investigative study.
\newblock In \emph{IEEE International Conference on Identity, Security and
  Behavior Analysis}, 2017.

\bibitem[Chen et~al.(2019{\natexlab{b}})Chen, Hui, Wang, Tsao, Shuai, and
  Cheng]{beautyglow}
Hung-Jen Chen, Ka-Ming Hui, Szu-Yu Wang, Li-Wu Tsao, Hong-Han Shuai, and
  Wen-Huang Cheng.
\newblock Beautyglow: On-demand makeup transfer framework with reversible
  generative network.
\newblock In \emph{Proceedings of the {IEEE} Conference on Computer Vision and
  Pattern Recognition}, 2019{\natexlab{b}}.

\bibitem[{Dantcheva} et~al.(2012){Dantcheva}, {Chen}, and
  {Ross}]{youtube_makeup_dataset}
A.~{Dantcheva}, C.~{Chen}, and A.~{Ross}.
\newblock Can facial cosmetics affect the matching accuracy of face recognition
  systems?
\newblock In \emph{IEEE International Conference on Biometrics: Theory,
  Applications and Systems}, 2012.

\bibitem[Deng et~al.(2019)Deng, Guo, Niannan, and Zafeiriou]{deng2018arcface}
Jiankang Deng, Jia Guo, Xue Niannan, and Stefanos Zafeiriou.
\newblock Arcface: Additive angular margin loss for deep face recognition.
\newblock In \emph{CVPR}, 2019.

\bibitem[Feng et~al.(2018)Feng, Wu, Shao, Wang, and Zhou]{feng2018prn}
Yao Feng, Fan Wu, Xiaohu Shao, Yanfeng Wang, and Xi~Zhou.
\newblock Joint 3d face reconstruction and dense alignment with position map
  regression network.
\newblock In \emph{Proceedings of the European Conference on Computer Vision},
  2018.

\bibitem[Gu et~al.(2019)Gu, Wang, Chiu, Tai, and Tang]{gu2019ladn}
Qiao Gu, Guanzhi Wang, Mang~Tik Chiu, Yu-Wing Tai, and Chi-Keung Tang.
\newblock Ladn: Local adversarial disentangling network for facial makeup and
  de-makeup.
\newblock In \emph{Proceedings of the International Conference on Computer
  Vision}, 2019.

\bibitem[Hertzmann et~al.(2001)Hertzmann, Jacobs, Oliver, Curless, and
  H.Salesin]{image_anatogy}
Aaron Hertzmann, Charles~E. Jacobs, Nuria Oliver, Brian Curless, and David
  H.Salesin.
\newblock Image analogies.
\newblock \emph{Proceedings of the ACM SIGGRAPH Conference on Computer
  Graphics}, 2001.

\bibitem[Jiang et~al.(2019)Jiang, Liu, Gao, Cao, He, Feng, and
  Yan]{jiang2019psgan}
Wentao Jiang, Si~Liu, Chen Gao, Jie Cao, Ran He, Jiashi Feng, and Shuicheng
  Yan.
\newblock Psgan: Pose and expression robust spatial-aware gan for customizable
  makeup transfer.
\newblock In \emph{Proceedings of the {IEEE} Conference on Computer Vision and
  Pattern Recognition}, 2019.

\bibitem[Kingma and Dhariwal(2018)]{glow}
Durk~P Kingma and Prafulla Dhariwal.
\newblock Glow: Generative flow with invertible 1x1 convolutions.
\newblock In \emph{Advances in Neural Information Processing Systems}, 2018.

\bibitem[Kips et~al.(2020)Kips, Gori, Perrot, and Bloch]{kips2020cagan}
Robin Kips, Pietro Gori, Matthieu Perrot, and Isabelle Bloch.
\newblock Ca-gan: Weakly supervised color aware gan for controllable makeup
  transfer.
\newblock In \emph{ECCV Workshops}, 2020.

\bibitem[{Li} et~al.(2015){Li}, {Zhou}, and {Lin}]{makeup_physic}
C.~{Li}, K.~{Zhou}, and S.~{Lin}.
\newblock Simulating makeup through physics-based manipulation of intrinsic
  image layers.
\newblock In \emph{Proceedings of the {IEEE} Conference on Computer Vision and
  Pattern Recognition}, 2015.

\bibitem[Li et~al.(2018)Li, Qian, Dong, Liu, Yan, Zhu, and Lin]{beautygan}
Tingting Li, Ruihe Qian, Chao Dong, Si~Liu, Qiong Yan, Wenwu Zhu, and Liang
  Lin.
\newblock Beautygan: Instance-level facial makeup transfer with deep generative
  adversarial network.
\newblock In \emph{Proceedings of the ACM international conference on
  Multimedia}, 2018.

\bibitem[Paysan et~al.(2009)Paysan, Knothe, Amberg, Romhani, and
  Vetter]{paysan09basel}
P~Paysan, R~Knothe, B~Amberg, S~Romhani, and T~Vetter.
\newblock A {3D} face model for pose and illumination invariant face
  recognition.
\newblock In \emph{Proceedings of the IEEE International Conference on Advanced
  Video and Signal Based Surveillance}, 2009.

\bibitem[Rathgeb et~al.(2019)Rathgeb, Dantcheva, and Busch]{overview_makeup}
Christian Rathgeb, Antitza Dantcheva, and Christoph Busch.
\newblock Impact and detection of facial beautification in face recognition: An
  overview.
\newblock \emph{IEEE Access}, 2019.

\bibitem[Richardson et~al.(2016)Richardson, Sela, and Kimmel]{richardson20163d}
Elad Richardson, Matan Sela, and Ron Kimmel.
\newblock 3d face reconstruction by learning from synthetic data.
\newblock In \emph{International Conference on 3D Vision}, 2016.

\bibitem[Romdhani and Vetter(2003)]{romdhani2003efficient}
Sami Romdhani and Thomas Vetter.
\newblock Efficient, robust and accurate fitting of a {3D} morphable model.
\newblock In \emph{Proceedings of the International Conference on Computer
  Vision}, 2003.

\bibitem[Tewari et~al.(2017)Tewari, Zollhofer, Kim, Garrido, Bernard, Perez,
  and Theobalt]{Tewari_2017_ICCV}
Ayush Tewari, Michael Zollhofer, Hyeongwoo Kim, Pablo Garrido, Florian Bernard,
  Patrick Perez, and Christian Theobalt.
\newblock {MoFA}: Model-based deep convolutional face autoencoder for
  unsupervised monocular reconstruction.
\newblock In \emph{Proceedings of the International Conference on Computer
  Vision}, 2017.

\bibitem[Tong et~al.(2017)Tong, Tang, Brown, and Xu]{example_comestic}
Wai-Shun Tong, Chi-Keung Tang, Michael~S Brown, and Ying-Qing Xu.
\newblock Example-based cosmetic transfer.
\newblock In \emph{Pacific Conference on Computer Graphics and Applications},
  2017.

\bibitem[Tran et~al.(2017)Tran, Hassner, Masi, and Medioni]{tran16_3dmm_cnn}
Anh Tran, Tal Hassner, Iacopo Masi, and G\'{e}rard Medioni.
\newblock Regressing robust and discriminative {3D} morphable models with a
  very deep neural network.
\newblock In \emph{Proceedings of the {IEEE} Conference on Computer Vision and
  Pattern Recognition}, 2017.

\bibitem[Tran et~al.(2018)Tran, Hassner, Masi, Paz, Nirkin, and
  Medioni]{tran2018extreme}
Anh~Tuan Tran, Tal Hassner, Iacopo Masi, Eran Paz, Yuval Nirkin, and
  G{\'e}rard~G Medioni.
\newblock Extreme 3d face reconstruction: Seeing through occlusions.
\newblock In \emph{Proceedings of the {IEEE} Conference on Computer Vision and
  Pattern Recognition}, 2018.

\bibitem[Tran and Liu(2018)]{tran2018nonlinear}
Luan Tran and Xiaoming Liu.
\newblock Nonlinear 3d face morphable model.
\newblock In \emph{Proceedings of the {IEEE} Conference on Computer Vision and
  Pattern Recognition}, 2018.

\bibitem[Vetter and Blanz(1998)]{vetter1998estimating}
Thomas Vetter and Volker Blanz.
\newblock Estimating coloured 3d face models from single images: An example
  based approach.
\newblock In \emph{Proceedings of the European Conference on Computer Vision},
  1998.

\bibitem[Wikipedia(2020)]{cosmeticWiki}
Wikipedia.
\newblock Cosmetic industry -- {Wikipedia}, 2020.
\newblock \url{https://en.wikipedia.org/w/index.php?title=Cosmetic_industry}.

\bibitem[Xu et~al.(2013)Xu, Du, and Zhang]{makeup_landmark}
Lin Xu, Yangzhou Du, and Yimin Zhang.
\newblock An automatic framework for example-based virtual makeup.
\newblock In \emph{Proceedings of the IEEE International Conference on Image
  Processing}, 2013.

\bibitem[Yang et~al.(2020)Yang, Zhu, Wang, Huang, Shen, Yang, and
  Cao]{yang2020facescape}
Haotian Yang, Hao Zhu, Yanru Wang, Mingkai Huang, Qiu Shen, Ruigang Yang, and
  Xun Cao.
\newblock Facescape: A large-scale high quality 3d face dataset and detailed
  riggable 3d face prediction.
\newblock In \emph{Proceedings of the {IEEE} Conference on Computer Vision and
  Pattern Recognition}, 2020.

\bibitem[Zhang et~al.(2019)Zhang, Chen, He, and Jin]{dmt}
Honglun Zhang, Wenqing Chen, Hao He, and Yaohui Jin.
\newblock Disentangled makeup transfer with generative adversarial network.
\newblock In \emph{arXiv}, 2019.

\bibitem[Zhang et~al.(2016)Zhang, Zhang, Li, and Qiao]{zhang2016joint}
Kaipeng Zhang, Zhanpeng Zhang, Zhifeng Li, and Yu~Qiao.
\newblock Joint face detection and alignment using multitask cascaded
  convolutional networks.
\newblock \emph{IEEE Signal Processing Letters}, 2016.

\bibitem[Zhu et~al.(2017)Zhu, Park, Isola, and Efros]{CycleGAN2017}
Jun-Yan Zhu, Taesung Park, Phillip Isola, and Alexei~A Efros.
\newblock Unpaired image-to-image translation using cycle-consistent
  adversarial networks.
\newblock In \emph{Proceedings of the International Conference on Computer
  Vision}, 2017.

\bibitem[Zhu et~al.(2016)Zhu, Lei, Liu, Shi, and Li.]{Zhu2016Face}
Xiangyu Zhu, Zhen Lei, Xiaoming Liu, Hailin Shi, and Stan~Z. Li.
\newblock Face alignment across large poses: A {3D} solution.
\newblock In \emph{Proceedings of the {IEEE} Conference on Computer Vision and
  Pattern Recognition}, 2016.

\end{thebibliography}
